\documentclass[review]{elsarticle}

\usepackage{lineno, hyperref}
\modulolinenumbers[5]
\usepackage[utf8]{inputenc} % allow utf-8 input
\usepackage[T1]{fontenc}    % use 8-bit T1 fonts
\usepackage{url}            % simple URL typesetting
\usepackage{booktabs}       % professional-quality tables
\usepackage{amsfonts}       % blackboard math symbols
\usepackage{nicefrac}       % compact symbols for 1/2, etc.
\usepackage{microtype}      % microtypography
\usepackage{graphicx}
\usepackage{pdflscape}
\usepackage[dvipsnames]{xcolor}
\usepackage{tikz}
\usetikzlibrary{bayesnet}
\usetikzlibrary{arrows}
\usepackage{makecell}
\usepackage{xspace}
\usepackage[ruled,vlined,algo2e]{algorithm2e}
\usepackage[noend]{algorithmic}
\usepackage{wrapfig}
\usepackage{subcaption}
\usepackage{xcolor}

\usepackage{bm}
\usepackage{color}
\usepackage{amssymb}

\usepackage{algorithmic}
\usepackage{algorithm}
\usepackage[cmex10]{amsmath}
\usepackage{url}

\usepackage{multirow}

\def\p{{\mathrm{p}}}
\def\q{{\mathrm{q}}}
\def\rd{{\mathrm{d}}}

\def\bmu{{\boldsymbol{\mu}}}

\def\bSigma{{\boldsymbol{\Sigma}}}

\def\bff{{\mathbf f}}

\def\bh{{\mathbf h}}

\def\bm{{\mathbf m}}

\def\bu{{\mathbf u}}

\def\bx{{\mathbf x}}
\def\by{{\mathbf y}}
\def\bz{{\mathbf z}}

\def\bC{{\mathbf C}}

\def\bI{{\mathbf I}}

\def\bK{{\mathbf K}}

\def\bS{{\mathbf S}}

\def\bX{{\mathbf X}}

\def\bZ{{\mathbf Z}}

\newcommand{\R}{\mathbb{R}}
\DeclareMathOperator*{\diag}{diag}

\begin{document}

\begin{frontmatter}

\title{Introducing instance label correlation in multiple instance learning. Application to cancer detection on histopathological images}

\author[UGR]{Pablo Morales-\'{A}lvarez\texorpdfstring{\fnref{visitUC}\corref{mycorrespondingauthor}}{}}\ead{pablomorales@decsai.ugr.es}
\author[UGR]{Arne Schmidt}
\author[UC]{José Miguel Hernández-Lobato}
\author[UGR]{Rafael Molina}

\address[UGR]{Department of Computer Science and AI, University of Granada, Spain}
\address[UC]{Department of Engineering, University of Cambridge, UK}

\fntext[visitUC]{Work done during the Margarita Salas fellowship outgoing phase at Cambridge University, UK.}
\cortext[mycorrespondingauthor]{Corresponding author}

\begin{abstract}

In the last years, the weakly supervised paradigm of multiple instance learning (MIL) has become very popular in many different areas.
A paradigmatic example is computational pathology, where the lack of patch-level labels for whole-slide images prevents the application of supervised models.
Probabilistic MIL methods based on Gaussian Processes (GPs) have obtained promising results due to their excellent uncertainty estimation capabilities. 
However, these are general-purpose MIL methods that do not take into account one important fact: in (histopathological) images, the labels of neighboring patches are expected to be correlated.
In this work, we extend a state-of-the-art GP-based MIL method, which is called VGPMIL-PR, to exploit such correlation. 
To do so, we develop a novel coupling term inspired by the statistical physics Ising model. 
We use variational inference to estimate all the model parameters. Interestingly, the VGPMIL-PR formulation is recovered when the weight that regulates the strength of the Ising term vanishes.
The performance of the proposed method is assessed in two real-world problems of prostate cancer detection.
We show that our model achieves better results than other state-of-the-art probabilistic MIL methods.
We also provide different visualizations and analysis to gain insights into the influence of the novel Ising term.
These insights are expected to facilitate the application of the proposed model to other research areas.

\end{abstract}


\end{frontmatter}

%\linenumbers

\section{Introduction}

Multiple instance learning (MIL) has caught great attention in fields where there is a challenging lack of labelled data. Although it has been applied in many different areas \cite{carbonneau2018multiple}, we will focus on the case of computational pathology.
In the last years, thanks to the increasing digitalization of whole-slide images (WSIs), the field of computational pathology is developing computer-aided diagnosis systems based on machine learning for cancer detection \cite{cui2021artificial, huang_bag_2022}.
% \cite{cui2021artificial, huss2020software}. 
The goal of computational pathology is to provide a fast and reliable diagnosis for the most prototypical cases, letting the pathologists focus on the most challenging ones. 
Ultimately, this will enable a much wider access to early cancer diagnosis \cite{niazi2019digital}.

% The problem of labelling in Computational Pathology
In order to make accurate predictions, machine learning classification methods need to be trained using a labelled set of instances \cite{murphy2012machine}. 
In the case of computational pathology, these instances are typically patches from the WSIs (and not the complete images themselves) \cite{lopez2021learning, arne_access, campanella2019clinical}.
% \cite{lopez2021learning, hou2016patch}. 
The reason for this is twofold: i) it is useful to have predictions at patch level in order to know where exactly in the image the cancer is located, and ii) WSIs are extremely large and cannot be directly fed to a classifier.
As a consequence, notice that expert pathologists must label \emph{every single patch} in the training data as cancerous or not (we will consider the binary problem cancer/no-cancer throughout this work). 
Given the large number of patches and the limited availability of pathologists, this becomes a daunting task in real practice \cite{schmidt2022coupling}.

% Weakly supervised learning - MIL
To address this problem, different weakly supervised learning paradigms have been proposed in recent years. 
% They all share the same philosophy, which is to reduce the labelling workload. 
Here we focus on MIL, which has become very popular in the medical domain \cite{wu2021combining, campanella2019clinical}.
The idea in MIL is that instances are grouped in bags, and only bag labels are needed for training.
In the case of WSIs, all the patches coming from the same image are considered a bag. 
Therefore, the labelling workload on pathologists decreases enormously: from labelling every single patch, to only labelling the complete WSI as cancerous or not.

% Need for probabilistic methods: GPs
Different machine learning algorithms have been developed to learn under the MIL setting.
Notice that dealing with uncertainty is essential in MIL models, since instance-level labels are unknown. 
To deal with uncertainties, different probabilistic methods have been developed, such as Dirichlet Process Mixture Models \cite{kandemir2014instance}, Markov chain \cite{read_multi-label_2017}, Monte-Carlo chain \cite{read_efficient_2014, read_probabilistic_2019} and Gaussian Processes (GPs) \cite{kim2010gaussian,BMVC2016_71}. In particular, GPs have attracted plenty of attention in the last years, due to their expressive power and their capacity to handle uncertainty in a principled manner. Moreover, we are interested in this type of probabilistic models, since they will allow for introducing correlations in a theoretically sound way.

%  VGPMIL and VGPMIL-Probit
Among GP-based MIL methods, we will focus on the two most successful ones: VGPMIL and VGPMIL-PR.
VGPMIL \cite{haussmann2017variational} was proposed in 2017 to overcome the limitations of two earlier formulations \cite{kim2010gaussian, BMVC2016_71} (namely, the use of the inefficient Laplace approximation and the impossibility to obtain instance-level predictions, respectively). In short, VGPMIL relies on variational inference and allows for closed-form updates of its parameters.
However, the use of the logistic function implies that VGPMIL needs to resort to a theoretical approximation during inference (namely, the Jaakola bound \cite[Eq. (10)]{haussmann2017variational}). As shown in \cite{wang2021multiple}, such approximation hurts predictive performance in practice. 
As an alternative, the authors of \cite{wang2021multiple} propose the utilization of the probit function, which removes the need for the aforementioned approximation. This method, which will be referred to as VGPMIL-PR, is considered the current state of the art among probabilistic MIL approaches.

% Patch correlation, smoothness conditions
Methods such as VGPMIL and VGPMIL-PR are general-purpose MIL models that can be used in any MIL problem (that is, whenever the label is known only at bag level, see different use-cases in \cite{haussmann2017variational, wang2021results}).
However, the underlying MIL assumption that the labels of the instances in a bag are independent of each other is unrealistic in many real problems. 
For example, in the particular case of WSI images (and in many image-related MIL problems), the labels of neighboring patches are expected to be correlated \cite{ding2015probabilistic}.
We hypothesize that the predictive performance of MIL methods can be enhanced by incorporating this type of prior knowledge into the model.

% In this work
In this work, we introduce a novel GP-based MIL algorithm that takes into account the correlation between the labels of neighboring patches, and we apply it to the real-world problem of prostate cancer detection on histopathological images. 
We model the correlation through a coupling term inspired by the Ising model \cite[Section 19.4.1]{murphy2012machine}, an statistical physics method that has found several applications in computer vision \cite{ding2015probabilistic, qin}.
Our GP-MIL modeling builds on VGPMIL-PR, so our method will be referred to as VGPMIL-PR-I (Ising).
In VGPMIL-PR-I, a hyperparameter $\lambda$ regulates the influence of the Ising-inspired terms.
Variational inference is used to estimate the model parameters, and the update formulas of VGPMIL-PR are recovered when $\lambda\to 0$ (that is, when the influence of the coupling term vanishes). 
In the experimental section, we show that VGPMIL-PR-I outperforms the state-of-the-art GP-based MIL approaches VGPMIL and VGPMIL-PR when predicting at both instance and bag levels, while keeping an analogous computational cost.
Moreover, to gain insights into the influence of the new coupling term, we analyze the role of $\lambda$, and provide several visualizations for the predictions. 

% Outline
The rest of the paper is organized as follows. Section \ref{sec:theory} presents the probabilistic model and inference for the novel VGPMIL-PR-I. Closely related methods such as VGPMIL and VGPMIL-PR are also discussed in this section.
Section \ref{sec:experiments} focuses on the empirical evaluation of the model, including the data description, the experimental framework, and the discussion of results.
Section \ref{sec:conclusions} provides the main conclusions and some future outlook.

\section{Probabilistic model and inference}\label{sec:theory}

In this section we present the theoretical description for VGPMIL-PR-I.
Specifically, Section \ref{sec:notation} explains the problem formulation and the main notation. 
Section \ref{sec:background} explains the closely-related methods VGPMIL and VGPMIL-PR, which are at the base of our formulation. 
Section \ref{sec:ising} introduces the novel coupling term that accounts for patch label correlation, which is used to define VGPMIL-PR-I.
Section \ref{sec:inference} shows how to perform variational inference to estimate the parameters in VGPMIL-PR-I. 
Section \ref{sec:predictions} explains the procedure to make predictions at both instance and bag levels.

\subsection{Notation and problem formulation}\label{sec:notation}

Our notation follows the state-of-the-art work \cite{wang21_MIL_probit}.
The training data is given by a set of bags $\bX=\{\bX_b\}_{b\in\mathcal{B}}$ and their corresponding labels $\by=\{y_b\}_{b\in\mathcal{B}}$. We deal with a binary problem, i.e. $y_b\in\{0,1\}$. Each bag $\bX_b=\{\bx_i\}_{i\in b}$ contains $|b|$ instances, i.e. $b=\{i_1,\dots,i_{|b|}\}\subseteq [N]$ ($N$ is the total amount of instances). Notice that different bags may have different amounts of instances.
Each instance $\bx_i$ is given by a vector in $\R^D$. 
In the MIL setting, one assumes that each instance has its (unknown) label $h_i\in\{0,1\}$.
We write $\bh_b$ for the labels of all the instances belonging to bag $b$.
The MIL labelling assumption dictates that a bag is considered positive (class 1) if at least one of its instances is positive. Mathematically, this is
\begin{equation}\label{eq:MIL_assumption}
\p(y_b|\bh_b)=1[y_b=\max_{i\in b} h_i],
\end{equation}
where $1[\cdot]$ is the indicator function (i.e. it equals one when its argument is true and zero otherwise). Finally, we will collectively denote $\bh=\{\bh_b\}_{b\in\mathcal{B}}$.

In the case of WSIs, each $\bX_b$ is an image, which is composed of its patches $\{\bx_i\}_{i\in b}$. Each patch has an unknown label $h_i$ (0 for non-cancerous and 1 for cancerous), and we only have access to the bag label $y_b$ (whether the image is cancerous or not, i.e. whether it contains at least one patch that is cancerous).

The goal in MIL is to train a model based only on bag labels $\{y_b\}_{b\in \mathcal{B}}$.
And such model must be able to predict at both instance and bag levels. That is, given a previously unseen instance $\bx^\star\in\R^D$, we are interested in the probability $\p(h^\star=1)$.
Likewise, given a previously unseen complete bag $\bX^\star$, we are interested in $\p(y^\star=1)$.

\subsection{Background: VGPMIL and VGPMIL-PR}\label{sec:background}

As mentioned in the introduction, our model is inspired by two closely related methods: VGPMIL and VGPMIL-PR.
To understand our contribution, it is essential to fully understand their formulations, which we explain next.

\subsubsection{VGPMIL formulation}\label{sec:vgpmil}

VGPMIL was introduced in \cite{haussmann2017variational}. 
The idea is to consider a sparse GP classification model \cite{snelson_sparse_2006} to describe the relationship between instance features $\bX$ and their (unknown) labels $\bh$. Then, an additional bag likelihood must be considered to model the (observed) bag labels $\by$ given the instance labels $\bh$. Both components are described next.

\textbf{The sparse GP classification model}. Instances $\bx_i$ are associated latent variables $f_i\in \R$ which are modelled through a GP, $f\sim \mathcal{GP}(0,\kappa)$. We write $\kappa$ for the GP kernel, which encodes the properties of the considered functions. 
Then, the instance labels $h_i$ are defined from $f_i$ through a classification likelihood $\nu$:
\begin{linenomath*}
\begin{equation}\label{eq:hgf}
\p(h_i|f_i)=\nu(f_i)^{h_i}(1-\nu(f_i))^{1-h_i}.    
\end{equation}
\end{linenomath*}
Specifically, VGPMIL uses the logistic function $\nu(x)=(1+e^{-x})^{-1}$.
Intuitively, a large (resp. low) value of $f_i$ implies that the class is likely to be one (resp. zero).  
Moreover, since standard GPs scale poorly with the number of training instances $N$, VGPMIL makes use of sparse GPs \cite{snelson_sparse_2006}, which summarize the training data through $M\ll N$ inducing points.
These inducing points $\bu=\{u_1,\dots,u_M\}$ represent the value of the GP at some inducing points locations $\bZ=\{\bz_1,\dots,\bz_M\}$ (just like $\bff=\{f_1,\dots,f_N\}$ are the GP values at $\bX=\{\bx_1,\dots,\bx_N\}$).
Therefore, the distributions of $\bu$ and $\bff|\bu$ are:
\begin{linenomath*}
\begin{align}
    \p(\bu) & = \mathcal{N}(\bu|\mathbf{0},\bK_{\bZ\bZ}), \label{eq:u} \\
    \label{eq:fgu}
    \p(\bff|\bu) &= \mathcal{N}(\bff|\bK_{\bX\bZ}\bK_{\bZ\bZ}^{-1}\bu,\mathcal{K}).
\end{align}
\end{linenomath*}
The expression of $\mathcal{K}$ is given by the particular sparse GP approach used in VGPMIL, which is FITC \cite{snelson_sparse_2006}, so we have $\mathcal{K}=\diag(\bK_{\bX\bX}-\bK_{\bX\bZ}\bK_{\bZ\bZ}^{-1}\bK_{\bZ\bX})$.
As is standard in GP literature, we are writing $\bK_{\bX\bX}$ for the $N\times N$ covariance matrix $\bK_{\bX\bX}=(\kappa(\bx_i,\bx_j))_{1\leq i,j\leq N}$. The definitions for $\bK_{\bX\bZ}$ and $\bK_{\bZ\bZ}$ are analogous.

\textbf{The bag likelihood}.
VGPMIL introduces the following parameterization to model the bag labels from the instance labels:
\begin{equation}\label{eq:ygh_vgpmil}
    \p(y_b|\bh_b)=\frac{H^{G_b}}{H+1},
\end{equation}
where $G_b:=1[y_b=\max_{i\in b} h_i]$ and $H$ is a large and fixed value (in their examples, they use $H=100$).
Eq.~\eqref{eq:ygh_vgpmil} approximates the MIL assumption introduced in eq.~\eqref{eq:MIL_assumption}: if some instance label $h_i$ is one, then the bag label $y_b$ is one with very high probability (namely, with probability $\frac{H}{H+1}$).
Otherwise (that is, if all instance labels $h_i$ are zero), the bag label is one with very low probability (namely, $\frac{1}{H+1}$).

In summary, VGPMIL is given by eqs.~\eqref{eq:u}, \eqref{eq:fgu}, \eqref{eq:hgf} and \eqref{eq:ygh_vgpmil}. For additional details, the interested reader is referred to the original work \cite{haussmann2017variational}.

\subsubsection{VGPMIL-PR formulation}\label{sec:vgpmil_pr}

\begin{figure}[t]
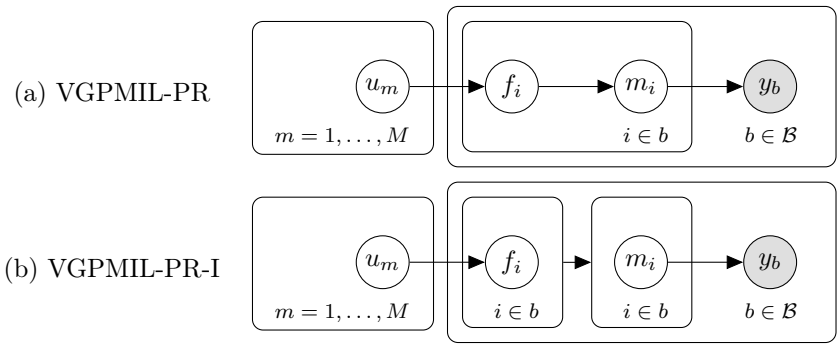

    \centering
    \begin{tabular}{cc}
    \multirow{1}{*}[2.5em]{(a) VGPMIL-PR}
    &
    \tikz{
    % nodes
     \node[latent] (u) {$u_m$};%
     \node[latent,right=of u] (f) {$f_i$}; %
     \node[latent,right=of f] (m) {$m_i$}; %
     \node[obs,right=of m] (y) {$y_b$}; %
    % plate
     \plate [inner sep=.3cm,xshift=.01cm,yshift=.2cm] {plate1} {(u)} {$m=1,\dots,M$}; %
     \plate [inner sep=.3cm,xshift=.01cm,yshift=.2cm] {plate2} {(f)(m)} {$i\in b$};
     \plate [inner sep=.5cm,xshift=.01cm,yshift=.2cm] {plate3} {(f)(m)(y)} {$b\in\mathcal{B}$};
    % edges
     \edge {u}{f}
     \edge {f}{m}
     \edge {m}{y}
     }
    \\
    \multirow{1}{*}[2.5em]{(b) VGPMIL-PR-I} &
    \tikz{
    % nodes
     \node[latent] (u) {$u_m$};%
     \node[latent,right=of u] (f) {$f_i$}; %
     \node[latent,right=of f] (m) {$m_i$}; %
     \node[obs,right=of m] (y) {$y_b$}; %
    % plate
     \plate [inner sep=.3cm,xshift=.01cm,yshift=.2cm] {plate1} {(u)} {$m=1,\dots,M$}; %
     \plate [inner sep=.3cm,xshift=.01cm,yshift=.2cm] {plate2} {(f)} {$i\in b$};
     \plate [inner sep=.3cm,xshift=.01cm,yshift=.2cm] {plate3} {(m)} {$i\in b$};
     \plate [inner sep=.5cm,xshift=.01cm,yshift=.2cm] {plate4} {(f)(m)(y)} {$b\in\mathcal{B}$};
    % edges
     \edge {u}{f}
     \edge {plate2}{plate3}
     \edge {m}{y}
     }
    \end{tabular}
    \caption{Probabilistic graphical model for VGPMIL-PR (a) and VGPMIL-PR-I (b). Gray nodes are observed variables, and white ones are latent variables to be estimated. The main difference is that VGPMIL-PR-I introduces correlation between instances in the same bag. Therefore, the distribution of $\bm_b$ given $\bff_b$ does not factorize across instances. The correlation is introduced through a novel term inspired by the Ising model, see Section \ref{sec:ising} for details.\label{fig:pgm}}
\end{figure}

VGPMIL-PR was recently proposed in \cite{wang2021multiple} as an improvement over VGPMIL.
Namely, the logistic function used by VGPMIL in eq.~\eqref{eq:hgf} is not conjugate with the Gaussian distribution coming from the GP, recall eqs.~\eqref{eq:u}--\eqref{eq:fgu}.
This means that, in order to achieve mathematical tractability, VGPMIL needs to resort to the Jaakola bound \cite[Eq. (10)]{haussmann2017variational}.
However, the use of this bound introduces an \emph{approximation} in the training objective. 
As shown in \cite{wang2021multiple}, such approximation damages the predictive performance in practice. 
Consequently, the authors of \cite{wang2021multiple} introduce an alternative formulation based on the probit function, VGPMIL-PR. They show that, via a variable augmentation approach, VGPMIL-PR allows for directly optimizing the training objective (without approximations).

More specifically, VGPMIL-PR uses the probit function $\nu(x)=\int_{-\infty}^x \mathcal{N}(t|0,1)\mathrm{d}t$ in eq.~\eqref{eq:hgf}.
Also, the bag likelihood is given by eq.~\eqref{eq:MIL_assumption} (instead of eq.~\eqref{eq:ygh_vgpmil}). 
Then, to circumvent the need for approximations, VGPMIL-PR leverages a variable augmentation approach \cite{girolami2006variational}.
Namely, for each instance we introduce a new variable $m_i\in\R$ between $f_i$ and $h_i$, which is defined as $m_i\sim\mathcal{N}(f_i,1)$. Since we are using a probit likelihood, we have that $h_i=1[m_i>0]$. Analogously to the rest of variables, we write $\bm_b=\{m_i\}_{i\in b}$ for all the $m_i$'s inside bag $b$, and we use $\bm=\{\bm_b\}_{b\in\mathcal{B}}$ to collectively denote all the $m_i$'s in the model. Then, by marginalizing out $\bh$, we have:
\begin{linenomath*}
\begin{align}
    \p(\bm|\bff) &= \prod_b \p(\bm_b|\bff_b) = \prod_b \mathcal{N}(\bm_b|\bff_b,I), \label{eq:mgf} \\
    \p(\by|\bm) & = \prod_b \p(y_b|\bm_b), \label{eq:ygm}
\end{align}
\end{linenomath*}
where $I$ is the identity matrix (of size $|b|$) and $\p(y_b=0|\bm_b)=\prod_{i\in b} 1[m_i<0]$. 
Importantly, these augmented variables $\bm$ will prove extremely helpful to introduce the Ising correlation in the next section.

In summary, VGPMIL-PR is given by eqs.~\eqref{eq:u}, \eqref{eq:fgu}, \eqref{eq:mgf} and \eqref{eq:ygm}. Figure \ref{fig:pgm}(a) shows the probabilistic graphical model for VGPMIL-PR.

% Comparisons to VGPMIL and VGPMIL-PR can go here or later, when our full model is specified. 

\subsection{Correlating patch labels: VGPMIL-PR-I}\label{sec:ising}

\begin{figure}[t]
    \centering
    \tikzset{every picture/.style={line width=0.75pt}} %set default line width to 0.75pt
\begin{tikzpicture}[x=0.75pt,y=0.75pt,yscale=-1,xscale=1]
%Shape: Square [id:dp3129525728260556] 
\draw   (130,73) -- (180,73) -- (180,123) -- (130,123) -- cycle ;
%Shape: Square [id:dp6156154448475408] 
\draw   (180,73) -- (230,73) -- (230,123) -- (180,123) -- cycle ;
%Shape: Square [id:dp17667792772400404] 
\draw   (130,123) -- (180,123) -- (180,173) -- (130,173) -- cycle ;
%Shape: Square [id:dp9763081415098203] 
\draw   (180,123) -- (230,123) -- (230,173) -- (180,173) -- cycle ;
%Shape: Square [id:dp5640271029877335] 
\draw   (230,123) -- (280,123) -- (280,173) -- (230,173) -- cycle ;

% Text Node
\draw (143,90) node [anchor=north west][inner sep=0.75pt]  [font=\Large]  {$P_{1}$};
% Text Node
\draw (193,90) node [anchor=north west][inner sep=0.75pt]  [font=\Large]  {$P_{2}$};
% Text Node
\draw (143,140) node [anchor=north west][inner sep=0.75pt]  [font=\Large]  {$P_{3}$};
% Text Node
\draw (193,140) node [anchor=north west][inner sep=0.75pt]  [font=\Large]  {$P_{4}$};
% Text Node
\draw (243,140) node [anchor=north west][inner sep=0.75pt]  [font=\Large]  {$P_{5}$};
\end{tikzpicture}
    \caption{A simplified illustration of an image composed by five patches: $P1,\dots,P5$.}
    \label{fig:ising_example}
\end{figure}
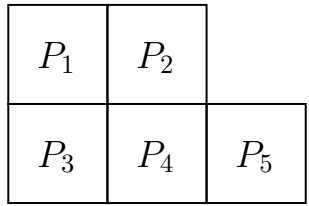

% VGPMIL and VGPMIL-PR as general methods
As mentioned in the introduction, VGPMIL and VGPMIL-PR are general MIL approaches that can be used in any MIL problem.
Indeed, there are plenty of applications where MIL methods can be used.
For instance, think of a recommendation system where a reviewer has not evaluated every single item in the database, but has reviewed ``groups'' of them (e.g., he/she likes science-fiction movies, although he/she has not rated individual movies).
Consider also a task of anomaly detection in which we do not have labels for individual transactions, but we only know whether there was some anomalous behavior in a certain period of time (which contains many different transactions).

% The particularity of images
Here we focus in the particular use-case of images, where bags are images and their instances are their patches.
In this case, there exists very valuable information coming from the structure of the image itself, which can be exploited in the model. 
For example, it is natural to think that neighboring patches in the same image are likely to have similar labels. 
The main goal of this work is to introduce such correlation into the VGPMIL-PR formulation. To do so, we are inspired by the Ising model. 

% Ising model in physics and ML
\textbf{The novel coupling term.} The Ising model arose from statistical physics to describe the behavior of magnets. 
In some magnets, called ferro-magnets, neighboring spins tend to line up in the same direction, whereas in other kinds of magnets,
called anti-ferromagnets, the spins are repelled from their neighbors \cite{murphy2012machine}. This type of interactions based on the Ising model have been used previously in machine learning and computer vision to describe relationships between pixels of an image, see e.g. \cite{ding2015probabilistic, qin, murphy2012machine}. However, to the best of our knowledge, they have never been used in the context of MIL. 

% Continuous counterpart (CAR)
Our first idea was to consider an Ising model over the patch labels of each image, $\{h_i\}_{i\in b}$. However, inference proved very challenging in this case, due to the non-conjugacy of the Ising model and the GP-based MIL formulation. 
As an alternative, we considered a continuous counterpart of the Ising model over the variables $\bm_b=\{m_i\}_{i\in b}$ introduced in VGPMIL-PR, which are directly related to the patch labels (recall from Section \ref{sec:vgpmil_pr} that $h_i=1[m_i>0]$). 
Notice that such continuous version of the Ising model corresponds to the well-known Conditional Autoregression (CAR) \cite{ripley2005spatial}.
Importantly, as we will see in the rest of this section, this alternative formulation yields a Gaussian distribution on $\bm_b$, which can be treated analytically together with the GP-based MIL model. 

% Ising model in our case
% In our case, we want to introduce correlation between contiguous patches inside an image. 
Specifically, we consider the following coupling term for each image, which is defined over the augmented variables $\bm_b$, recall Section \ref{sec:vgpmil_pr}:
\begin{linenomath*}
\begin{align}\label{eq:coupling_term}
    \mathcal{C}(\bm_b)=&\exp\left(-\frac{\lambda}{2}\cdot
\sum_{\substack{i,j\in b\\ i<j}}\mathbf{1}[i,j\textrm{ are contiguous}]\cdot(m_i-m_j)^2
\right)=\nonumber \\
=&\exp\left(-\frac{\lambda}{2}\bm_b^\intercal \bC_b \bm_b\right).
\end{align}
\end{linenomath*}
Notice that $\mathcal{C}(\bm_b)$ is always in the range $[0,1]$, and it becomes close to zero when the value of $m$ is very different for neighboring patches.
For the second equality in eq.~\eqref{eq:coupling_term}, notice that the sum only produces quadratic terms in $m$, so it can be written as $\bm_b^\intercal \bC_b \bm_b$ for some positive semidefinite matrix $\bC_b$.

Since $m$ determines the label of each patch (recall from Section \ref{sec:vgpmil_pr} that $h_i=1[m_i>0]$), the term $\mathcal{C}(\bm_b)$ can be used to favor ``smoothness'' in the labels associated to the different patches. Also, the hyperparameter $\lambda$, which can be set to any non-negative value, regulates the strength of the coupling term: the larger $\lambda$, the more importance is given to differences in $m$. For example, when $\lambda=0$, $\mathcal{C}(\bm_b)$ becomes constant and it does not account for correlation between patch labels.

% Illustrating the Ising term 
\textbf{An example of $\mathcal{C}(\bm_b)$.}
To illustrate the proposed coupling term, consider an image with five patches $P_1,\dots,P_5$ distributed as in Figure \ref{fig:ising_example}.
In this case, the quadratic terms of $\bm_b$ are:
\begin{linenomath*}
\begin{equation}
    (m_1-m_2)^2+(m_1-m_3)^2+(m_2-m_4)^2+(m_3-m_4)^2+(m_4-m_5)^2,
\end{equation}
\end{linenomath*}
and therefore we have:
\begin{linenomath*}
\begin{equation}\label{eq:Cb_example}
    \bC_b = 
    \begin{pmatrix}
 2 & -1 & -1 & 0 & 0 \\
 -1 & 2 & 0 & -1 & 0 \\
-1 & 0 & 2 & -1 & 0 \\
0 & -1 & -1 & 3 & -1\\
0 & 0 & 0 & -1 & 1 \\
\end{pmatrix}.
\end{equation}
\end{linenomath*}
In general, it is easy to compute the matrix $\bC_b$ for any given image. Notice that it is always a positive semidefinite matrix, and thus it is associated to a (singular) normal distribution.

% Introducing the term in the equations (and recovering the previous one)
\textbf{The VGPMIL-PR-I formulation}.
To introduce the new coupling term $\mathcal{C}(\bm_b)$ in the MIL formulation, we modify eq.~\eqref{eq:mgf} and define:
\begin{linenomath*}
\begin{equation}
\p(\bm|\bff) \propto \prod_b \mathcal{C}(\bm_b)\cdot\mathcal{N}(\bm_b|\bff_b,I).
\end{equation}
\end{linenomath*}
Notice that the probability of a configuration $\bm_b$ is proportional to the coupling term $\mathcal{C}(\bm_b)$, which favors smoothness across labels of neighboring patches. 
Decisively, since both $\mathcal{C}(\bm_b)$ and $\mathcal{N}(\bm_b|\bff_b,I)$ only contain (the exponential of) quadratic terms in $\bm_b$, the new distribution can be written as a Gaussian: 
\begin{linenomath*}
\begin{equation}\label{eq:mgf_ising}
    \p(\bm|\bff) = \prod_b \mathcal{N}(\bm_b|\bSigma_b\bff_b, \bSigma_b),
\end{equation}
\end{linenomath*}
with $\bSigma_b = (\lambda \bC_b+I)^{-1}$. Notice that this new formulation provides a generalization of VGPMIL-PR. Namely, when $\lambda\to 0$, we have that $\bSigma_b \to I$ and we recover eq.~\eqref{eq:mgf}.

In summary, the proposed model is given by eqs.~\eqref{eq:u}, \eqref{eq:fgu}, \eqref{eq:mgf_ising}, and \eqref{eq:ygm}. 
Notice also that, instead of FITC, in eq.~\eqref{eq:fgu} we leverage the more recent sparse GP approach introduced in \cite{hensman2015scalable}. Basically, this means that $\mathcal{K}=\bK_{\bX\bX}-\bK_{\bX\bZ}\bK_{\bZ\bZ}^{-1}\bK_{\bZ\bX}$ in eq.~\eqref{eq:fgu}.
Our method will be referred to as VGPMIL-PR-I (I denotes Ising). The probabilistic graphical model is depicted in Figure \ref{fig:pgm}(b).

\subsection{Variational inference}\label{sec:inference}

% Mean-field VI
In order to make inference in the proposed model, we need to compute the posterior distribution $\p(\bu,\bff,\bm|\by)$.
However, this is not analytically tractable due to the definition of the bag likelihood in eq.~\eqref{eq:ygm}, which depends on the sign of the $m_i$'s.
Following \cite{haussmann2017variational} and \cite{wang2021multiple}, we leverage standard mean-field variational inference (VI) theory \cite[Section 10.1.1]{bishop} to calculate an approximate posterior distribution that factorizes as
\begin{linenomath*}
\begin{equation}
    \q(\bu,\bff,\bm) = \q(\bu)\p(\bff|\bu)\q(\bm).
\end{equation}
\end{linenomath*}
Applying the well-known mean-field VI update equation \cite[Eq. (10.9)]{bishop}, we have that $\q(\bu)$ and $\q(\bm)$ can be iteratively computed as
\begin{linenomath*}
\begin{align}
    \q(\bu)&=\mathcal{N}(\bu|\bmu^u,\bSigma^u),
    \label{eq:qu_update}
    \\
    \label{eq:qm_update}
    \q(\bm)&\propto \prod_b\p(y_b|\bm_b)\mathcal{N}(\bm_b|\bmu^{\bm_b},\bSigma_b),
\end{align}
\end{linenomath*}
where
\begin{linenomath*}
\begin{align}
    \bSigma^u 
    &=
    \left( 
    \bK_{\bZ\bZ}^{-1}+
    \bK_{\bZ\bZ}^{-1}\bK_{\bZ\bX}\bSigma\bK_{\bX\bZ}\bK_{\bZ\bZ}^{-1}
    \right)^{-1},
    \label{eq:sigma_u}
    \\
    \label{eq:mu_u}
    \bmu^u &= \bSigma^u\bK_{\bZ\bZ}^{-1}\bK_{\bZ\bX}\mathbb{E}_{\q(\bm)}(\bm),
\end{align}
\end{linenomath*}
and 
\begin{linenomath*}
\begin{equation}\label{eq:mu_mb}
    \bmu^{\bm_b}=\bSigma_b\bK_{b\bZ}\bK_{\bZ\bZ}^{-1}\bmu^u.
\end{equation}
\end{linenomath*}
Here we are writing $\bSigma$ for the $N\times N$ block-diagonal matrix that contains all the $\bSigma_b$'s, $b\in\mathcal{B}$. Also, we are writing $\bK_{b\bZ}$ for the $|b|\times M$ matrix of covariances between $\bX_b$ and $\bZ$. 
Very importantly, notice that these update rules generalize those derived in \cite{wang2021multiple} for VGPMIL-PR.
Namely, when $\lambda\to 0$, we have that $\bSigma_b, \bSigma \to I$, and then eqs.~\eqref{eq:qu_update}--\eqref{eq:mu_mb} match eqs.(15)--(19) in \cite{wang2021multiple}. 

% Analyzing the required computations
All the computations involved in eqs.~\eqref{eq:qu_update}--\eqref{eq:mu_mb} are straightforward, except for $\mathbb{E}_{\q(\bm)}(\bm)$.
Indeed, each $\q(\bm_b)$ is a multivariate Gaussian truncated to $(-\infty,0)^{|b|}$ (or $\R^{|b|}\setminus(-\infty,0)^{|b|}$, depending on whether $y_b=0$ or $y_b=1$, respectively). 
It is well-known that the expectation of a truncated multivariate Gaussian cannot be obtained in closed-form \cite{wilhelm2010tmvtnorm}. 
Notice that this is not an issue for VGPMIL-PR \cite{wang2021multiple}, where the absence of Ising terms implies dealing with \emph{univariate} Gaussians, whose expectations can be analytically computed. 

To overcome the problem, we first tried to leverage numerical methods proposed in \cite{li2015efficient} to approximate the expectation for the multivariate truncated case.
However, these methods proved computationally too expensive to be integrated within our iterative calculation of $\q(\bu)$ and $\q(\bm)$.
Therefore, we decided to approximate the multivariate Gaussian $\mathcal{N}(\bm_b|\bmu^{\bm_b},\bSigma_b)$ by the factorized $\mathcal{N}(\bm_b|\bmu^{\bm_b},\mathrm{diag}(\bSigma_b))$ and utilize the expression for one-dimensional truncated Gaussians.
Although such approximation reduces the influence of the Ising correlation at this specific computation, notice that the coupling terms, which are included in $\bSigma_b$, affect the update equations in more places across eqs.~\eqref{eq:sigma_u}--\eqref{eq:mu_mb}.

% Details on how to compute E_{q(m)}(m)
Specifically, the expression for each $\mathbb{E}_{\q(\bm_b)}(\bm_b)$ is as follows. 
For bags with $y_b=0$, we have that each $\q(m_i)$, $i\in b$, is a univariate normal distribution $\mathcal{N}((\bmu^{\bm_b})_i,(\bSigma_b)_{ii})$ truncated to $(-\infty,0)$.
The expectation of such a distribution is well-known and can be obtained in closed-form \cite{johnson1995continuous}:
\begin{linenomath*}
\begin{equation}\label{eq:qmi_0}
     E_i = \mu_i-\frac{\phi(\mu_i/\sigma_i)}{1-\Phi(\mu_i/\sigma_i)}\sigma_i,
\end{equation}
\end{linenomath*}
where $\phi$ and $\Phi$ are, respectively, the density and cumulative distribution functions of a standard Gaussian $\mathcal{N}(0,1)$ (recall that both are efficiently implemented in standard software packages such as Python's Scipy). We have also abbreviated $\mu_i=(\bmu^{\bm_b})_i$ and $\sigma_i=\sqrt{(\bSigma_b)_{ii}}$.
For bags with $y_b=1$, we proceed analogously to \cite{wang2021results} to obtain the normalization constant $Z$ of the distribution of interest (that is, the factorized Gaussian $\mathcal{N}(\bm_b|\bmu^{\bm_b},\mathrm{diag}(\bSigma_b))$ truncated to $\mathbb{R}^{|b|}-(-\infty,0)^{|b|}$).  
Then, the expectation of each $\q(m_i)$, $i\in b$, is given by: 
\begin{linenomath*}
\begin{equation}\label{eq:qmi_1}
    \mathbb{E}_{\q(m_i)}(m_i) = \frac{\mu_i - (1-Z)E_i}{Z},
\end{equation}
\end{linenomath*}
where $Z=1-\prod_{i\in b} (1-\Phi(\mu_i/\sigma_i))$, $E_i$ is given by eq.~\eqref{eq:qmi_0}, and we are again abbreviating $\mu_i=(\bmu^{\bm_b})_i$ and $\sigma_i=\sqrt{(\bSigma_b)_{ii}}$.

% Summary and algorithm
The full training algorithm is summarized in Algorithm \ref{alg}.
It is an iterative process that alternates the updates between $\q(\bu)$ and $\q(\bm)$.
Details on the GP kernel and initializations used in this work are provided in Section \ref{sec:exp_framework}.
The code for the proposed method will be publicly available upon acceptance of the paper.

\begin{algorithm}[t]
% \small
\SetKwInOut{Input}{Input}
\SetKwInOut{Output}{Output}

\Input{Bags $\bX=\{\bX_b\}_{b\in\mathcal{B}}$ and bag labels $\by=\{y_b\}_{b\in\mathcal{B}}$.}

Calculate the matrices $\bC_b$ that account for the instance correlation inside each bag $b\in\mathcal{B}$, recall eq.~\eqref{eq:coupling_term} and the example at eq.~\eqref{eq:Cb_example}.

Initialize GP kernel parameters and inducing points locations, as well as the posterior distributions $\q(\bu)$ and $\q(\bm)$. Details on initializations in the text.

\ForEach{$\mathrm{iteration}\ t=1,\dots,T$}{

Update $\q(\bu)$, using eqs.~\eqref{eq:qu_update}, \eqref{eq:sigma_u} and \eqref{eq:mu_u}.

Update $\q(\bm)$ and obtain $\mathbb{E}_{\q(\bm)}(\bm)$, using eqs.~\eqref{eq:qm_update}, \eqref{eq:mu_mb}, \eqref{eq:qmi_0}, and \eqref{eq:qmi_1}.

}
\Output{Posterior distributions $\q(\bu)$ and $\q(\bm)$.}

\caption{Training procedure for VGPMIL-PR-I. \label{alg}}

\end{algorithm}

\subsection{Making predictions}\label{sec:predictions}
Suppose we are given a new bag $\bX^\star=\{\bx^\star_i\}_{i \in b_\star}$.
As explained at the end of section \ref{sec:notation}, we are interested in both instance-level and bag-level predictions.
For this, we first need to compute the predictive distributions over $\bm^\star$.

By using the learned posterior $\q(\bu)$ along with $\p(\bff|\bu)$, we can obtain the joint distribution over $\bff^\star$:
\begin{linenomath*}
\begin{equation}
    \p(\bff^\star)=\int \p(\bff^\star|\bu)\p(\bu)\rd\bu=
    \mathcal{N}(\bff^\star|\bmu^\star, \bS^\star),
\end{equation}
\end{linenomath*}
with $\bmu^\star$ and $\bS^\star$ given by the standard sparse GP predictions:
\begin{linenomath*}
\begin{equation}\label{eq:f_star}
    \bmu^\star=\bK_{\star \bZ}\bK_{\bZ\bZ}^{-1}\bmu^u,\quad
    \bS^\star=\bK_{\star\star}-\bK_{\star \bZ}\bK_{\bZ\bZ}^{-1}(\bK_{\bZ\bZ}-\bSigma^u)\bK_{\bZ\bZ}^{-1}\bK_{\bZ\star}.
\end{equation}
\end{linenomath*}
Here, $\bmu^u$ and $\bSigma^u$ are the parameters learned during training, recall eq.~\eqref{eq:sigma_u} and \eqref{eq:mu_u}. Naturally, the subscript $\star$ in the kernel matrices $\bK$ indicates that we are using the new bag $\bX^\star$.
Then, since the distribution $\p(\bm|\bff)$ is also Gaussian, recall eq.~\eqref{eq:mgf_ising}, we can compute the joint distribution over $\bm^\star$ in closed-form:
\begin{linenomath*}
\begin{equation}
    \p(\bm^\star)=\int \p(\bm^\star|\bff^\star)\p(\bff^\star)\rd\bff^\star=
    \mathcal{N}(\bm^\star|\bmu_m^\star, \bS_m^\star),
\end{equation}
\end{linenomath*}
with $\bmu_m^\star$ and $\bS_m^\star$ given by
\begin{linenomath*}
\begin{equation}\label{eq:m_star}
    \bmu_m^\star=\bSigma_\star\cdot\bmu^\star,\quad
    \bS_m^\star=\bSigma_\star+\bSigma_\star\cdot\bS^\star\cdot\bSigma_\star^\intercal.
\end{equation}
\end{linenomath*}
Here, $\bmu^\star$ and $\bS^\star$ are given by eq.~\eqref{eq:f_star}, and $\bSigma_\star$ is the matrix that accounts for correlation among instances in the test bag $\bX^\star$, which is defined analogously to the training case, recall matrix $\bSigma_b$ in eq.~\eqref{eq:mgf_ising}.

Once we have the joint distribution over $\bm^\star$, the instance-level and bag-level predictions are given as:
\begin{linenomath*}
\begin{align}
    \p(h_i^\star=1)&=\p(m_i^\star>0)=
    \Phi\left((\bmu_m^\star)_i/\sqrt{(\bS_m^\star)_{ii}}\right),
    \label{eq:pred_instance}
    \\
    \label{eq:pred_bag}
    \p(y^\star=1)&=1-\int_{\bm^\star \in (-\infty,0)^{|b_\star|}}\p(\bm^\star)\rd\bm^\star.
\end{align}
\end{linenomath*}
Notice that the integral in eq.~\eqref{eq:pred_bag} can be computed efficiently with the cumulative distribution function of a multivariate Gaussian, which is also available in most standard statistical packages, such as Python's Scipy.

Interestingly, these predictions generalize those obtained in VGPMIL-PR \cite{wang2021multiple}. Indeed, if we do not consider correlation among instances in the test bag, i.e. $\bSigma_\star=\bI$, then eqs.~\eqref{eq:pred_instance} and \eqref{eq:pred_bag} match those in \cite{wang2021multiple} (last two equations before Section 3.5). 
Finally, although here we have detailed how to make predictions for a complete previously unseen bag $\bX^\star$, the same process can be applied to make predictions on previously unseen individual instances $\bx^\star$ (patches).

\section{Experiments}\label{sec:experiments}

In this section we thoroughly evaluate VGPMIL-PR-I in a real-world problem of prostate cancer detection. 
The experimental framework, including data, metrics and baselines, is explained in Section \ref{sec:exp_framework}.
The results are discussed in Section \ref{sec:results}.
Finally, in Section \ref{sec:panda} we evaluate our method in a much larger prostate cancer detection dataset: the well-known PANDA challenge.

\begin{figure}[t]
    \centering
    \begin{tabular}{cccc}
    \includegraphics[width=0.23\textwidth]{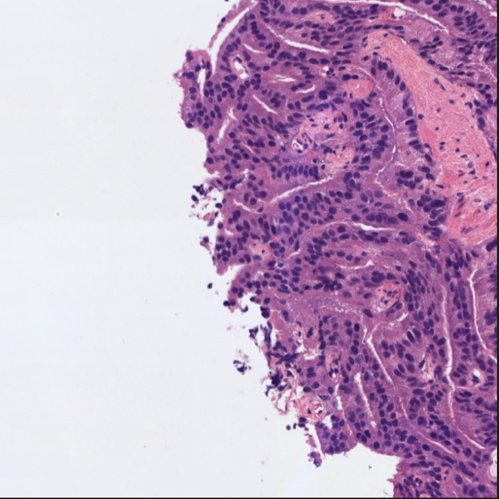} &
    \includegraphics[width=0.23\textwidth]{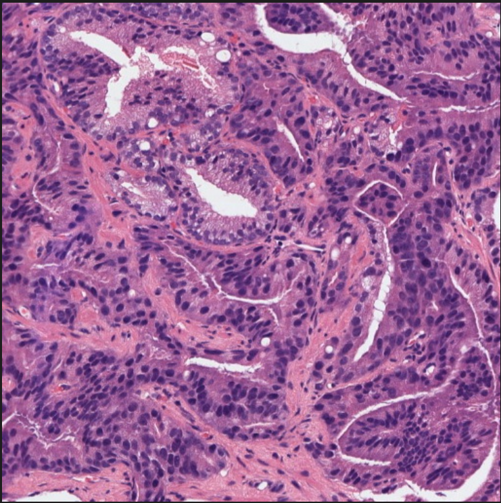} &
    \includegraphics[width=0.23\textwidth]{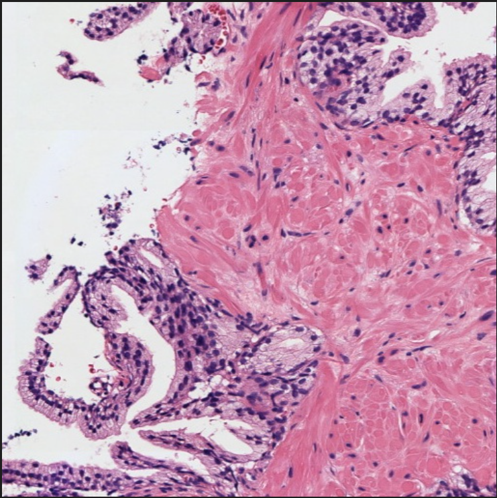} &
    \includegraphics[width=0.23\textwidth]{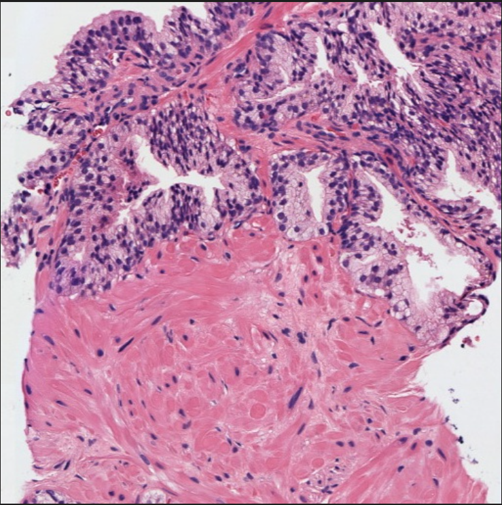} 
    \\
    \multicolumn{2}{c}{Cancerous} & 
    \multicolumn{2}{c}{Non-cancerous}
    \end{tabular}
    \caption{Two examples of cancerous (left) and non-cancerous (right) patches in the SICAPv2 test set.}
    \label{fig:patches_examples}
\end{figure}

\subsection{Experimental framework}\label{sec:exp_framework}

% Motivation for the problem
\textbf{Data description}.
In this paper we focus on the problem of prostate cancer detection.
However, notice that the algorithm can be applied for any other type of cancer (and more generally, for any other type of image).
Prostate cancer is the most commonly occurring cancer in men, and the second most commonly occurring cancer overall, according to the latest 2020 statistics on age-standardized incidence rate from the World Health Organisation (WHO) Global Cancer Observatory \cite{gco_who}.
We will use the prostate cancer database presented in \cite{silva2020going}, which is called SICAPv2 and is publicly available.
Although this database includes information on the Gleason score, which is used to evaluate the severity of the disease, in this work we will focus on the binary task of presence/absence of cancer.

% Data statistics
We use the original partition of the dataset, which contains 95 training and 31 test WSIs, respectively.
These very large images are split in 512x512 patches, resulting in a total amount of 15132 patches for training and 5246 for testing. Following the MIL paradigm, for the training set we only use binary labels benign/malign at the level of images (bags), but we do not have information at the level of patches (instances). 
In order to evaluate the predictive performance at instance-level, we do have labels for the patches in the test set. 
The amount of cancerous (resp. non-cancerous) images for the train set is 70 (resp. 25). For the test set, it is 25 (resp. 6). For illustration purposes, a couple of cancerous and non-cancerous patches are shown in Figure \ref{fig:patches_examples}. 
In order to train our model, each patch is represented through a 128-dimensional feature vector extracted in previous work \cite{arne_access}.

% Baselines (including details) and metrics.
\textbf{Baselines and metrics}.
Since our model is framed in the field of probabilistic GP-based MIL methods, we compare with the two most popular approaches VGPMIL \cite{haussmann2017variational} and VGPMIL-PR \cite{wang2021multiple}, which were reviewed in Section \ref{sec:background}.
For a fair comparison, the parameters used for the baselines are the same as those used for our method (described in next paragraph). 
For those parameters that do not have an analogous in our method (e.g. the initialization of $\q(\by)$ in VGPMIL), we use the default values proposed in the original papers.
To evaluate the performance of the compared methods we use four metrics: accuracy, precision, recall and F1-score (which provides a trade-off between precision and recall). 
For all the metrics, we use the standard implementations in the popular Python scikit-learn library \cite{scikit-learn}.

% Experimental details: kernel used, number of inducing points, parameters in the algorithm, number of iterations T, value of \lambda
\textbf{Experimental details}.
For the underlying GPs, in this work we use the well-known squared exponential kernel \cite{williams2006gaussian}, i.e. $\kappa(\bx,\by)=\gamma\cdot\exp\left(-||\bx-\by||^2/(2\ell^2)\right)$.
Following \cite{haussmann2017variational} and \cite{wang2021multiple}, we use standard values for the kernel hyperparameters, i.e. $\gamma=1$ and $\ell$ equals the square root of the number of features of $\bx, \by$ (in this work we set $\ell=11\approx \sqrt{128}$).
The number of inducing points is set to $M=200$, and their locations are initialized through K-means clustering as in previous work \cite{haussmann2017variational, wang2021multiple} (namely, $100$ of them are obtained by doing clustering on the patches that belong to the positive images, and the other $100$ on the patches that belong to the negative ones).
The number of iterations is set to $T=200$, which was enough to achieve convergence in practice. The expectation of the posterior distribution $\mathbb{E}_{\q(\bm)}(\bm)$ is initialized with a standard Gaussian for each instance independently. 
Notice that the initialization of $\q(\bu)$ is irrelevant since it gets updated first in Algorithm \ref{alg}. 
As for the value of $\lambda$, which regulates the strength of the Ising correlation (recall eq.~\eqref{eq:coupling_term}), we will analyze five different values in the experiments, $\lambda\in\{0.1,0.5,1.0,5.0,10.0\}$.
This will allow us to empirically illustrate the effect of $\lambda$.

\subsection{Experimental results}\label{sec:results}

\begin{table}[t]
    \setlength{\tabcolsep}{5pt}
    \centering
    \small
    \begin{tabular}{cccccc}
\toprule
                     & $\lambda$   & Accuracy & Precision &    Recall &   F1-score \\
\midrule
VGPMIL  & -      &       92.22$\pm$0.00 &        96.40$\pm$0.00 &     92.29$\pm$0.00 &     94.30$\pm$0.00 \\
VGPMIL-PR & - &       92.38$\pm$0.03 &        96.44$\pm$0.06 &     92.48$\pm$0.06 &     94.42$\pm$0.02 \\
\midrule
\multirow{5}{*}[-0.2em]{VGPMIL-PR-I} & 0.1  &       92.94$\pm$0.05 &        96.24$\pm$0.15 &     93.52$\pm$0.12 &     94.86$\pm$0.04 \\
                     & \textbf{0.5}  &       \textbf{93.85$\pm$0.04} &        \textbf{97.17$\pm$0.19} &     \textbf{93.90$\pm$0.14} &     \textbf{95.51$\pm$0.02} \\
                     & \textbf{1.0}  &       \textbf{94.58$\pm$0.03} &        \textbf{97.32$\pm$0.06} &     \textbf{94.83$\pm$0.04} &     \textbf{96.06$\pm$0.02} \\
                     & \textbf{5.0}  &       \textbf{95.11$\pm$0.06} &        \textbf{97.74$\pm$0.14} &     \textbf{95.18$\pm$0.09} &     \textbf{96.44$\pm$0.04} \\
                     & \textbf{10.0} &       \textbf{95.03$\pm$0.14} &        \textbf{97.72$\pm$0.16} &     \textbf{95.09$\pm$0.06} &     \textbf{96.39$\pm$0.10} \\
\bottomrule
\end{tabular}
    \caption{Predictive performance at the level of patches (instances). In bold, we highlight the values of $\lambda$ for which VGPMIL-PR-I gets better (or equal) performance than both baselines in all the metrics. The results are the mean and standard deviation over five independent runs.}
    \label{tab:instance_results}
\end{table}

\begin{table}[t]
    \setlength{\tabcolsep}{5pt}
    \centering
    \small
    \begin{tabular}{cccccc}
\toprule
                     & $\lambda$   & Accuracy & Precision &    Recall &   F1-score \\
\midrule
VGPMIL    & -    &  83.87$\pm$0.00 &   83.33$\pm$0.00 &  100.00$\pm$0.00 &  90.91$\pm$0.00 \\
VGPMIL-PR & - &  90.32$\pm$0.00 &   89.29$\pm$0.00 &  100.00$\pm$0.00 &  94.34$\pm$0.00 \\
\midrule
\multirow{5}{*}[-0.2em]{VGPMIL-PR-I}
 & \textbf{0.1}  &  \textbf{93.55$\pm$0.00} &   \textbf{92.59$\pm$0.00} &  \textbf{100.00$\pm$0.00} &  \textbf{96.15$\pm$0.00} \\
                     & \textbf{0.5}  &  \textbf{93.55$\pm$0.00} &   \textbf{92.59$\pm$0.00} &  \textbf{100.00$\pm$0.00} &  \textbf{96.15$\pm$0.00} \\
                     & 1.0  &  90.32$\pm$0.00 &   92.31$\pm$0.00 &   96.00$\pm$0.00 &  94.12$\pm$0.00 \\
                     & 5.0  &  87.10$\pm$0.00 &   95.65$\pm$0.00 &   88.00$\pm$0.00 &  91.67$\pm$0.00 \\
                     & 10.0 &  83.87$\pm$0.00 &   95.45$\pm$0.00 &   84.00$\pm$0.00 &  89.36$\pm$0.00 \\
\bottomrule
\end{tabular}
    \caption{Predictive performance at the level of images (bags). In bold, we highlight the values of $\lambda$ for which VGPMIL-PR-I gets better (or equal) performance than both baselines in all the metrics. The results are the mean and standard deviation over five independent runs.}
    \label{tab:bag_results}
\end{table}

\begin{table}[t]
    \centering
    \setlength{\tabcolsep}{5pt}
    \begin{tabular}{cc|cc|cc}
\multicolumn{1}{c}{} &\multicolumn{1}{c}{} &\multicolumn{2}{c}{VGPMIL} &\multicolumn{2}{c}{VGPMIL-PR} \\ 
\multicolumn{1}{c}{} & 
\multicolumn{1}{c|}{} & 
\multicolumn{1}{c}{Neg.} & 
\multicolumn{1}{c|}{Pos.} &
\multicolumn{1}{c}{Neg.} & 
\multicolumn{1}{c}{Pos.}
\\ \hline
\multirow[c]{2}{*}{\rotatebox[origin=tr]{90}{Actual}}
& Neg.  & 1 & 5  & 3 & 3 \\[1.5ex]
& Pos.  & 0   & 25 & 0   & 25  \\ \hline
\end{tabular}

\vspace{5mm}

    \begin{tabular}{cc|cc|cc|cc|cc|cc}
\multicolumn{1}{c}{} & \multicolumn{1}{c}{} & \multicolumn{10}{c}{VGPMIL-PR-I} 
\\
\multicolumn{1}{c}{} &\multicolumn{1}{c}{} &\multicolumn{2}{c}{$\lambda=0.1$} &\multicolumn{2}{c}{$\lambda=0.5$} 
&\multicolumn{2}{c}{$\lambda=1$}
&\multicolumn{2}{c}{$\lambda=5$}
&\multicolumn{2}{c}{$\lambda=10$}\\ 
\multicolumn{1}{c}{} & 
\multicolumn{1}{c|}{} & 
\multicolumn{1}{c}{Neg.} & 
\multicolumn{1}{c|}{Pos.} &
\multicolumn{1}{c}{Neg.} & 
\multicolumn{1}{c|}{Pos.} &
\multicolumn{1}{c}{Neg.} & 
\multicolumn{1}{c|}{Pos.} &
\multicolumn{1}{c}{Neg.} & 
\multicolumn{1}{c|}{Pos.} &
\multicolumn{1}{c}{Neg.} & 
\multicolumn{1}{c}{Pos.}
\\ \hline
\multirow[c]{2}{*}{\rotatebox[origin=tr]{90}{Actual}}
& Neg.  & 4 & 2& 4 & 2& 4 & 2& 5 & 1& 5 & 1 \\[1.5ex]
& Pos.   & 0   & 25& 0   & 25& 1   & 24& 3   & 22& 4   & 21   \\ \hline
\end{tabular}
    \caption{Confusion matrices obtained at the level of images for the compared methods.}
    \label{tab:cms}
\end{table}

\begin{figure}[t]
    \makebox[\textwidth][c]{\includegraphics[scale=0.43]{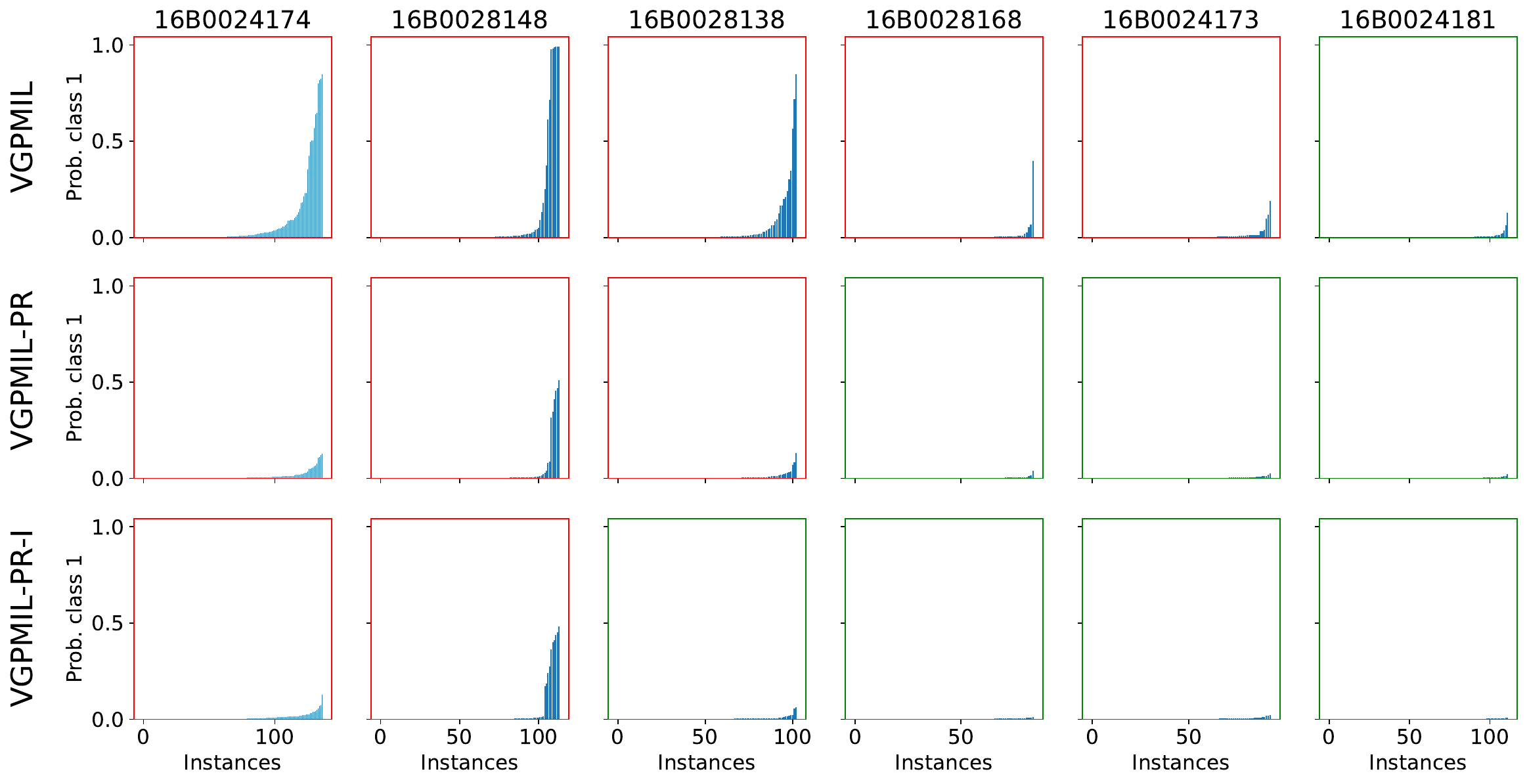}}
    \caption{Patch-level predictions inside each one of the six negative (non-cancerous) WSIs in the test set. Each column is an image (the header is the image identifier in the SICAPv2 dataset). The rows refers to the three compared methods.
    Each subplot has red/green axis depending on whether the image is correctly classified or not by that method. 
    The blue bars inside the subplots represent the probability of cancer for the different patches inside the image (for ease of visualization, they are sorted in increasing order). 
    }
    \label{fig:negative_bags}
\end{figure}

\begin{figure}%[t]
    \centering
    \begin{tabular}{c|c}
        \includegraphics[width=0.45\textwidth]{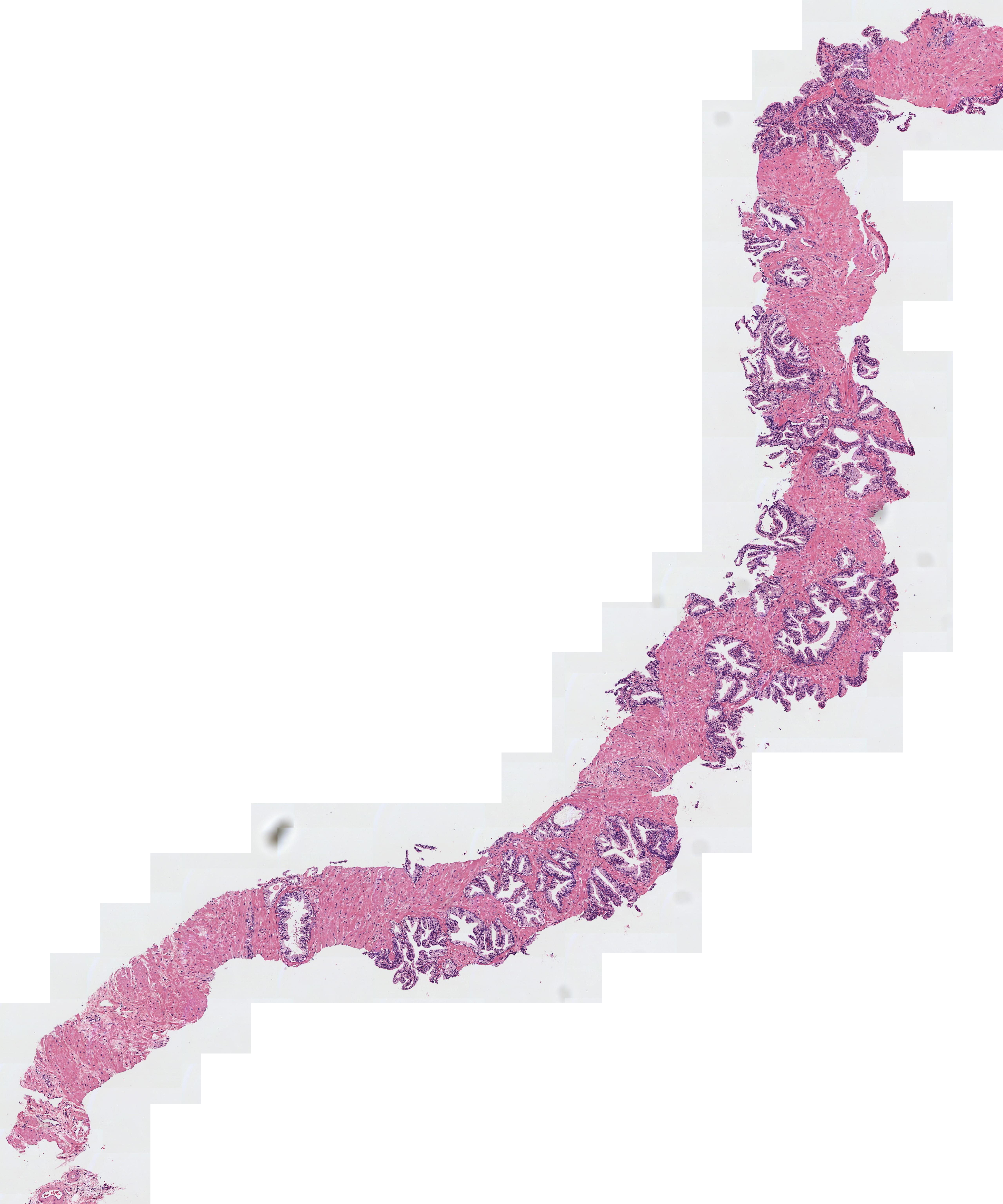} &
        \includegraphics[width=0.45\textwidth]{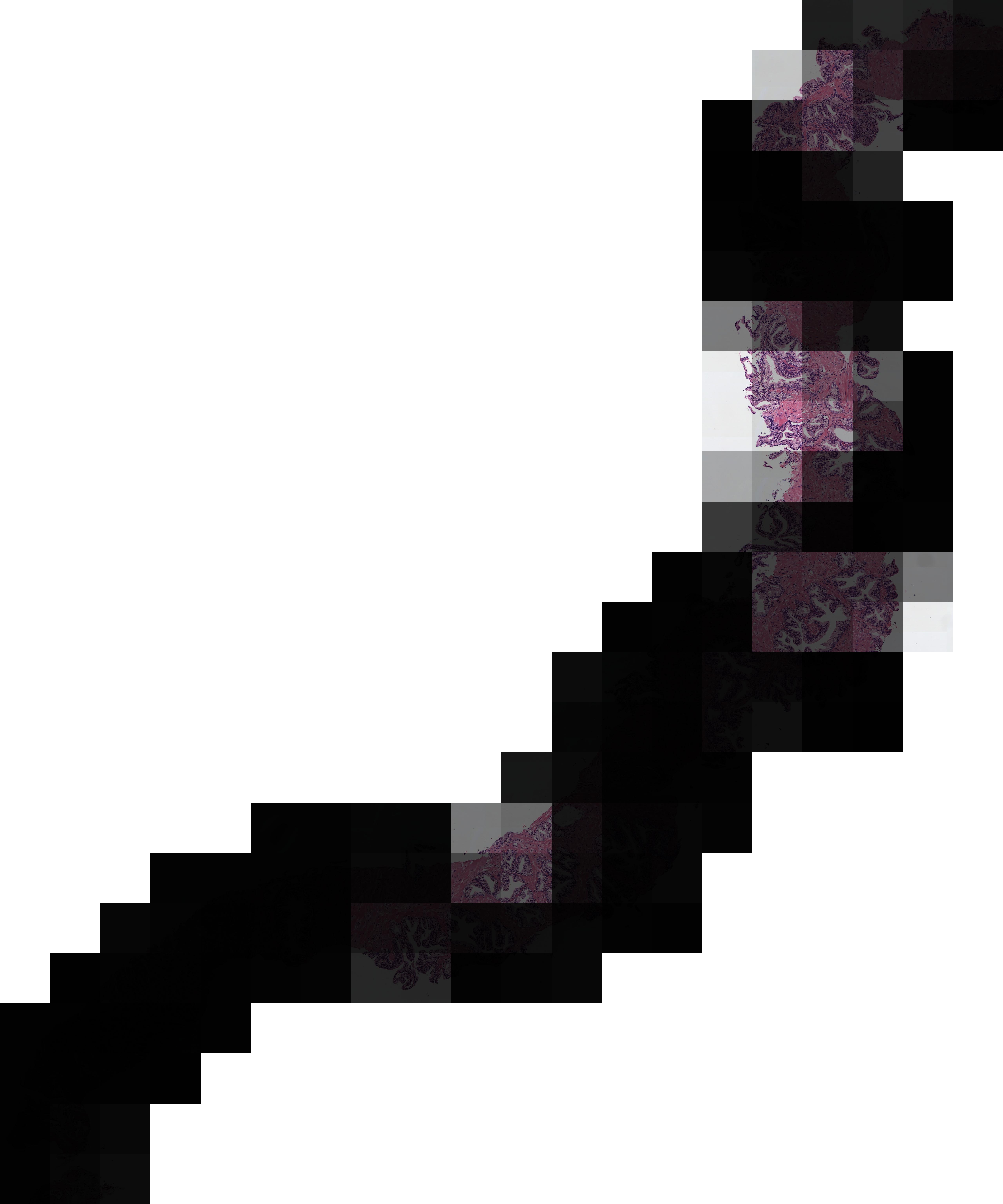}
        \\
        (a) Original image & (b) VGPMIL predictions \\
        \hline
        \\[-2mm]
        \includegraphics[width=0.45\textwidth]{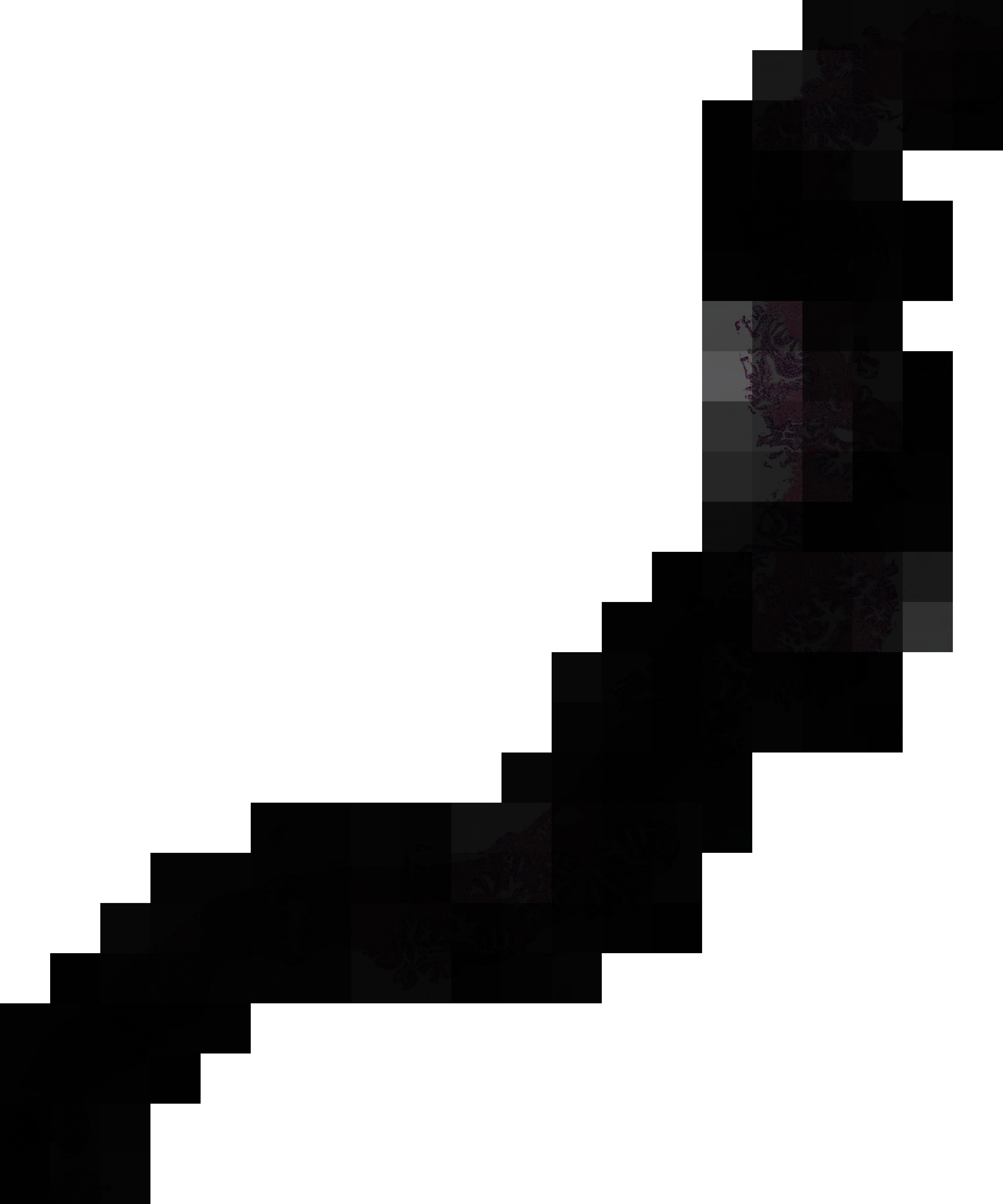} & \includegraphics[width=0.45\textwidth]{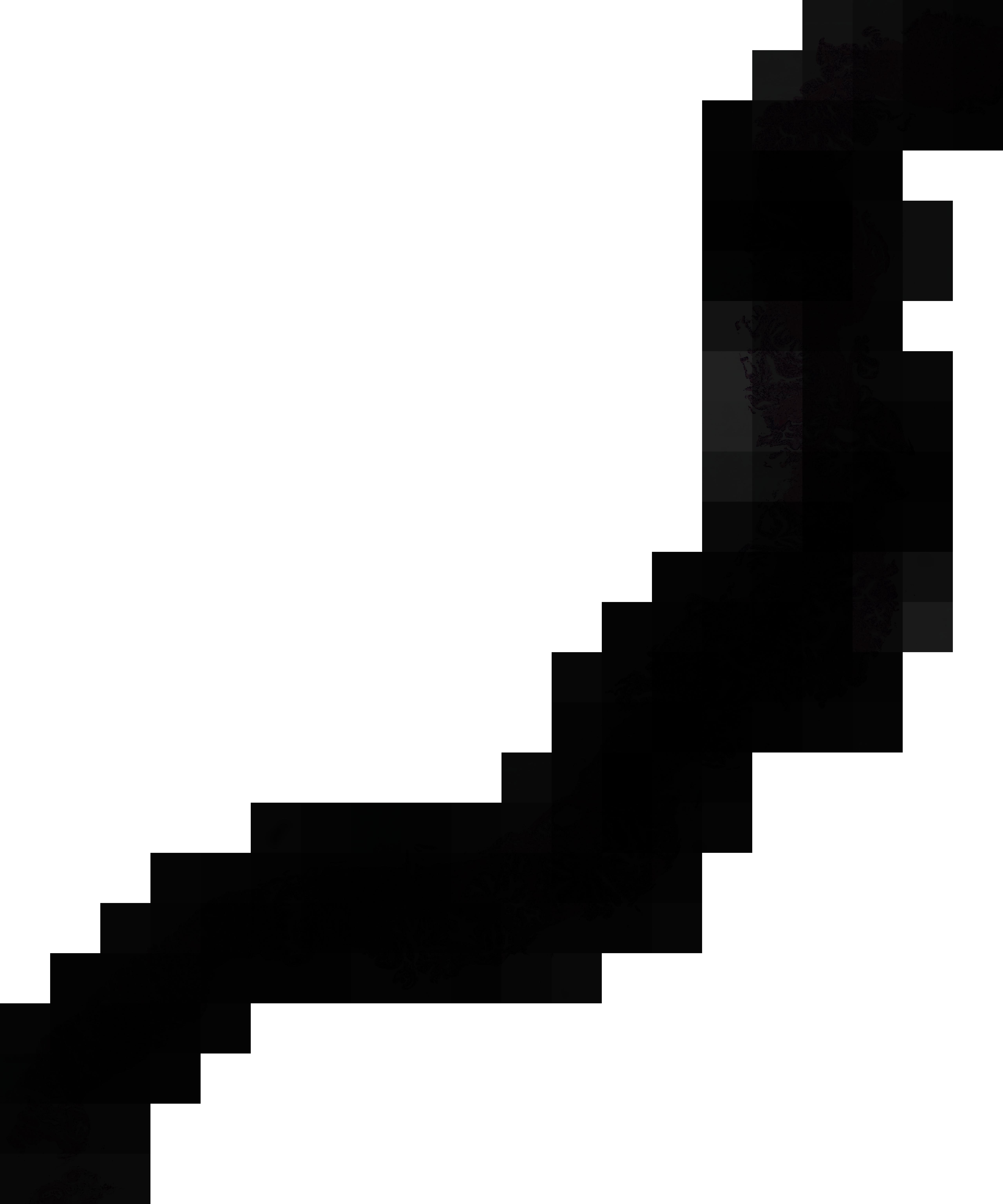} \\
        (c) VGPMIL-PR predictions & (d) VGPMIL-PR-I predictions
    \end{tabular}
    \caption{Patch level predictions obtained by the compared methods for image 16B0028138, which is non-cancerous. The original image is shown in (a). For predictions (b)--(d), the brightness of the patch is proportional to the probability of cancer (the brighter, the more probability).}
    \label{fig:visual_preds}
\end{figure}

\begin{figure}%[t]
    \centering
    \includegraphics[width=0.5\textwidth]{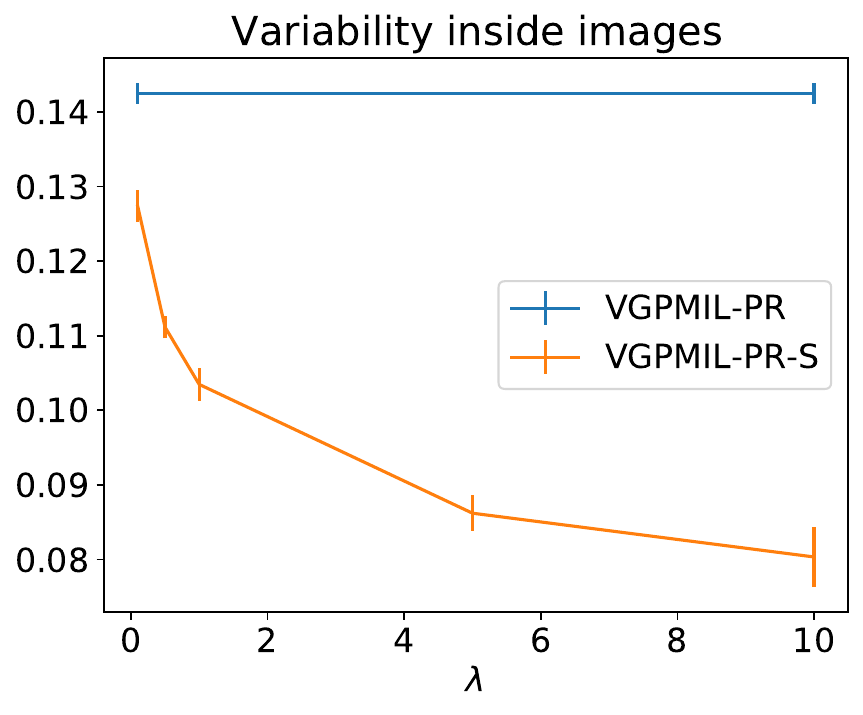}
    \caption{Variability in the patch-level predictions inside images. In VGPMIL-PR-I, the variability decreases as the strength of the Ising terms, given by $\lambda$, increases. VGPMIL-PR, which does not include Ising terms (i.e. $\lambda=0$), gets larger variability. As explained in the text, the variability inside an image is measured as the standard deviation of the probability of cancer for all the patches inside that image.}
    \label{fig:lambda_explicit}
\end{figure}

\begin{table}[t]
    \centering
    \begin{tabular}{cccc}
\toprule
            &   $\lambda$    &  Training time &        Testing time \\
\midrule
VGPMIL & -  & 15.45$\pm$0.53 &  3.14$\pm$0.16 \\
VGPMIL-PR & -  &  11.78$\pm$0.75 &  2.63$\pm$0.12 \\
\midrule
\multirow{5}{*}[-0.2em]{VGPMIL-PR-I} &  0.1  & 12.21$\pm$0.26  &  2.23$\pm$0.09 \\
            &  0.5  & 12.18$\pm$0.57  &  2.21$\pm$0.07 \\
            &  1.0  & 11.79$\pm$0.44  &  2.24$\pm$0.15 \\
            &  5.0  & 11.80$\pm$0.74  &  2.22$\pm$0.08 \\
            &  10.0 &  11.64$\pm$0.91 &  2.26$\pm$0.09 \\
\bottomrule
\end{tabular}
    \caption{Computational cost for training and testing the compared methods (in seconds). We are using 200 iterations in all cases, recall the experimental details in Section \ref{sec:exp_framework}. The results are the mean and standard deviation over five independent runs.\label{tab:times}}
\end{table}

In this section we evaluate the performance of the compared methods on the aforementioned
prostate cancer problem. We analyze eight different research questions, which are
discussed in the following paragraphs.

\vspace{1mm}
\noindent\textbf{Predictions at the level of instances (patches)}.
Although they only use bag labels for training, the compared methods can make predictions at the level of instances, recall Section \ref{sec:predictions}.
This is important to determine more precisely in which region (patch) the cancer is present.
Table \ref{tab:instance_results} shows the results when making predictions at patch level. 
We observe that VGPMIL-PR-I outperforms both baselines in all the metrics for four out of five values of $\lambda$ (and for $\lambda=0.1$, the baselines are only better in terms of precision).   
We also appreciate that VGPMIL-PR-I results are robust across different runs, obtaining low values of standard deviation. This stability is important for real-world applications, where one wants to avoid high sensitivity to random initializations. Finally, notice that, as argued in \cite{wang2021results}, we also observe that VGPMIL-PR (slightly) outperforms VGPMIL.

\vspace{1mm}
\noindent\textbf{Predictions at the level of bags (images)}.
Table \ref{tab:bag_results} shows the results when making predictions at the level of images.
We observe that VGPMIL-PR-I outperforms both baselines in all the metrics when $\lambda\in\{0.1,0.5\}$.
However, when $\lambda$ becomes larger, the results of VGPMIL-PR-I get worse. 
This fact can be explained theoretically because, whereas having low-to-moderate correlation among patches can be helpful, having strong ones tends to make the predictions too homogeneous, damaging the bag-level prediction (which takes into account the correlation among patches).
Indeed, in the next research question we analyze with greater detail how $\lambda$ is affecting the predictions on cancerous and non-cancerous images separately, which will provide additional insights.
Finally, similar to the patch-level results, we observe that VGPMIL-PR obtains better results than VGPMIL, as expected. Also, the predictions at the level of images are very stable across runs (notice the zero standard deviation).

\vspace{1mm}
\noindent\textbf{Analyzing the confusion matrices at bag level}.
Here we analyze more in detail the results presented in the previous paragraph (i.e. at the level of images). Table \ref{tab:cms} shows the confusion matrices for the compared methods.
Notice that both baselines classify all the 25 positive (cancerous) images correctly.
However, the difficulties arise at the non-cancerous images. 
This happens because of the nature of the MIL problem: as soon as a few patches obtain a non-negligible probability of cancer, the image will be likely predicted as cancerous (recall that the MIL formulation establishes that a bag has positive class as soon as one instance inside the bag has positive class).

% Interestingly, the correlation introduced by the Ising model can help at this, since they contribute to keep patch-level predictions more homogeneous.
Interestingly, the Ising model can help to avoid isolated false positive predictions on the patch-level (which lead to false positive bag predictions) by using the patch-level correlation.
This is reflected in Table \ref{tab:cms}, where we observe that increasingly more negative images are predicted correctly as $\lambda$ gets higher.
In contrast, notice that strong correlation damage the performance in the positive class, as they penalize the appearance of positive patches (which would break the homogeneity of the bag, where most patches do not contain cancer). 
Therefore, we conclude that instance correlation is beneficial when used with a low-to-medium intensity. For instance, in this application $\lambda=0.5$ is the best performing value, and it will be the one used by default in the sequel.

\vspace{1mm}
\noindent\textbf{Analyzing the instance-level predictions for negative bags}.
In the previous paragraph, we have explained that negative images are incorrectly classified because a few patches inside them get classified as positive. Here we provide a visualization to support this. Figure \ref{fig:negative_bags} shows how the patch-level predictions are distributed inside the six negative images available in the test set. We observe that the amount of patches with a non-negligible probability of cancer gets reduced as we move from VGPMIL to VGPMIL-PR, and then to VGPMIL-PR-I. 
This translates into better performance at bag-level (observe that the amount of green-axis subplots increases in the same sequence VGPMIL $\rightarrow$ VGPMIL-PR $\rightarrow$ VGPMIL-PR-I).
Notice that the improvement from VGPMIL to VGPMIL-PR is larger than from VGPMIL-PR to VGPMIL-PR-I.
This may be due to the simplification that was introduced when computing the expectation of the truncated multivariate Gaussian in VGPMIL-PR-I, recall the third-to-last paragraph in Section \ref{sec:inference}.

\vspace{1mm}
\noindent\textbf{Visualizing the predictions}.
In the last paragraph, we have analyzed how the patch-level predictions are distributed inside non-cancerous images \emph{quantitatively}. 
Indeed, Figure \ref{fig:negative_bags} represents each patch through a bar.
However, this hampers the \emph{qualitative} visualization from a medical viewpoint. 
Here we focus on such qualitative assessment by visualizing the predictions obtained for image 16B0028138, see Figure \ref{fig:visual_preds}. 
We have chosen this image because it illustrates best the effect of the coupling term.
Notice that, thanks to these terms, the proposed VGPMIL-PR-I manages to keep all patches with a probability closer to zero than VGPMIL and VGPMIL-PR. 
As a consequence, VGPMIL-PR-I is the only method that correctly classifies this image as non-cancerous, recall Figure \ref{fig:negative_bags}.
For the other two methods, there are some patches that trigger the image prediction to be cancerous.

\vspace{1mm}
\noindent\textbf{An explicit analysis on the role of $\lambda$}.
The hyperparameter $\lambda$ is at the core of the novel VGPMIL-PR-I.
It was introduced in the probabilistic model to regulate the strength of the coupling term, recall eq.~\eqref{eq:coupling_term}.
This role has been confirmed \emph{indirectly} in Table \ref{tab:cms}: when $\lambda$ gets higher, the predictive performance improves for negative bags and degrades for positive ones. 
This can be explained because a higher $\lambda$ homogenizes the patch-level predictions and difficulties the appearance of positive patches. 
Here we perform a more \emph{direct} measure to gain insights into the role of $\lambda$.
Specifically, we define the ``variability inside a bag'' as the standard deviation of the probability of cancer for all the patches inside that bag. Therefore, this metric measures the dispersion in the patch-level predictions obtained inside a bag.
Figure \ref{fig:lambda_explicit} shows the evolution of this metric for VPGMIL-PR-I as $\lambda$ increases.
%for VPGMIL-PR-I in our experiments.
As theoretically expected, the metric decreases as $\lambda$ gets higher. 
Also, notice that the metric value for VGPMIL-PR is higher. This is explained because VGPMIL-PR does not incorporate Ising correlation, i.e. $\lambda=0$. The value for VPGMIL is even higher, $0.30$, and it is not included in Figure \ref{fig:lambda_explicit} for ease of visualization. This greater value is probably due to the additional approximations that VGPMIL involves, which deepens the independence among patches.  

\vspace{1mm}
\noindent\textbf{Computational cost}.
Finally, we report the computational training and testing time for the compared methods, see Table \ref{tab:times}.
The results are in the same order of magnitude in all cases, which justifies the practical utility of the novel VGPMIL-PR-I, which obtained better predictive performance, recall Tables \ref{tab:instance_results} and \ref{tab:bag_results}.
In fact, VGPMIL-PR-I is slightly faster than VGPMIL, since the Jaakola bound approximation leveraged in the latter introduces additional parameters $\xi$ to be estimated.
As theoretically expected, the computational cost of VGPMIL-PR-I and VGPMIL-PR is analogous, since the update equations for the former are just a generalization of those for the latter, recall Sections \ref{sec:inference} and \ref{sec:predictions}.
Finally, notice that the value of $\lambda$ does not affect the computational cost of VGPMIL-PR-I, as $\lambda$ only regulates the intensity of the Ising terms (but it does not introduce any additional computation).

\vspace{1mm}
\noindent\textbf{Comparison to other related MIL approaches}.
So far we have focused on the comparison of VGPMIL-PR-I with VGPMIL and VGPMIL-PR. 
Since VGPMIL-PR-I builds on the same type of GP-based modeling, this is the most meaningful comparison in order to evaluate our main contribution (the Ising term to account for correlations among patches).
However, to provide a wider perspective, it is interesting to compare the novel VGPMIL-PR-I to other state-of-the-art and popular families of MIL methods. We consider three families: attention-based methods, where the two algorithms proposed in \cite{ilse2018attention} are the most popular approaches; MIL methods based on pseudo-labels such as the recent \cite{arne_access}; and classical pooling/aggregation methods such us the mean aggregation \cite{zhu2017deep}. These will be referred to as Att-MIL, Gated-Att-MIL, PS-MIL and Mean-Agg, respectively. 

Let us discuss the results both at instance (patch) and bag (image) levels. 
For the former, notice that the formulation of attention-based methods (Att-MIL and Gated-Att-MIL) and classical aggregation methods (Mean-Agg) do not allow for making predictions at instance level in a natural way. 
Namely, the instance-level labels are not modelled explicitly in this type of methods, and this is precisely one of their main limitations \cite{arne_agp_2023}.
Compared to PS-MIL, which does model instance labels explicitly, the novel VGPMIL-PR-I achieves higher predictive performance (95.11 vs 85.01 in accuracy and 96.44 vs 88.05 in F1-Score). Regarding bag-level performance, the results are shown in Table \ref{tab:att_pseudo_mean}. 
We observe that the best results are obtained by attention-based methods Att-MIL and Gated-Att-MIL, followed by the novel VGPMIL-PR-I. 

In conclusion, we observe that the results are quite different depending on the nature of the model and the user requirements. 
If one is interested in predictions at patch level, then the novel VGPMIL-PR-I is the best choice.
However, if one is only interested in image level performance, then attention-based approaches are the best option for this data.
Indeed, we hypothesise that the performance of attention-based approaches could be even enhanced by leveraging correlation among patches in a similar way to VGPMIL-PR-I. This is a interesting line of future research, see Section \ref{sec:conclusions}.

\begin{table}[t]
    \setlength{\tabcolsep}{5pt}
    \centering
    \small
    \begin{tabular}{rcccc}
\toprule
                     &  Accuracy & Precision &    Recall &   F1-score \\
\midrule

     VGPMIL-PR-I       &  93.55$\pm$0.00 &   92.59$\pm$0.00 &  \textbf{100.00}$\pm$0.00 &  96.15$\pm$0.00 \\
\midrule
Att-MIL & \textbf{96.80}$\pm$0.00 & \textbf{96.20}$\pm$0.00 & \textbf{100.00}$\pm$0.00 & \textbf{98.00}$\pm$0.00 \\
Gated-Att-MIL & \textbf{96.80}$\pm$ 0.00 & \textbf{96.20}$\pm$0.00 & \textbf{100.00}$\pm$0.00 & \textbf{98.00}$\pm$0.00 \\
Mean-Agg & 87.10$\pm$0.00 & 95.70$\pm$0.00 & 88.00$\pm$0.00 & 91.70$\pm$0.00 \\
PS-MIL & 90.32$\pm$NA & 89.28$\pm$NA & \textbf{100.00}$\pm$NA& 94.33$\pm$NA \\
\bottomrule
\end{tabular}
    \caption{Comparison with other related MIL methods which are not based on the GP modeling. The predictive performance at the level of images (bags) is shown. The results are the mean and standard deviation over five independent runs. The algorithm PS-MIL was run only once because of its high computational training cost.}
    \label{tab:att_pseudo_mean}
\end{table}

\subsection{Evaluation on a larger dataset: PANDA}\label{sec:panda}
The SICAPv2 dataset used so far is of medium-size (total amount of 126 WSIs, leading to 20378 patches; recall Section \ref{sec:exp_framework}). 
This has allowed us to carry out a very detailed analysis of the results.
In this section we show that the novel VGPMIL-PR-I also performs well on larger datasets, such us the well-known PANDA set. 
Although scalability is not an issue from a theoretical perspective, since the model is based on sparse GPs, it is important to
verify it in practice.

PANDA also tackles the problem of prostate cancer detection, and was presented at the MICCAI 2020 conference as a challenge\footnote{\url{https://panda.grand-challenge.org/}}.
Since the test set of PANDA is not publicly available, we use
the train/test split proposed in \cite{silva2021self}, where each split follows the overall class proportions. Namely, the dataset used here features a total amount of 10503 WSIs, which leads to 1107931 patches.
Notice that this is much larger than SICAPv2 (83 times larger in terms of WSIs).

Table \ref{tab:panda_bag} shows the predictive performance at image level, an aspect where attention-based methods stood out in the previous dataset. 
In this case, we observe that VGPMIL-PR-I obtains consistently better results. 
Additionally, as outlined in the last research question in Section \ref{sec:results}, VGPMIL-PR-I is able to provide instance-level predictions, which is not the case for attention-based models. We conclude that, for the PANDA dataset, the proposed method is the best choice in comparison to the other tested approaches.

\begin{table}[t]
    \setlength{\tabcolsep}{5pt}
    \centering
    \small
    \begin{tabular}{rcccc}
\toprule
                     &  Accuracy & Precision &    Recall &   F1-score \\
\midrule

     VGPMIL       &  74.87$\pm$0.08 &   74.25$\pm$0.06 &  \textbf{99.77$\pm$0.00} &  85.13$\pm$0.04 \\
     VGPMIL-PR       &  90.64$\pm$0.06 &   90.97$\pm$0.05 &  96.60$\pm$0.04 &  93.70$\pm$0.04 \\
     VGPMIL-PR-I       &  \textbf{92.57$\pm$0.14} &   \textbf{95.42$\pm$0.24} &  94.22$\pm$0.11 &  \textbf{94.82$\pm$0.09} \\
\midrule
Att-MIL & 90.92$\pm$0.01 & 94.17$\pm$0.01 & 93.43$\pm$0.01 & 93.79$\pm$0.01 \\
Gated-Att-MIL & 91.70$\pm$0.01 & 94.31$\pm$0.01 & 94.40$\pm$0.01 & 94.34$\pm$0.01 \\
Mean-Agg & 88.09$\pm$0.00 & 91.57$\pm$0.01 & 92.27$\pm$0.01 & 91.91$\pm$0.00 \\
PS-MIL & 88.36$\pm$NA & 87.99$\pm$NA & 97.11$\pm$NA & 92.33$\pm$NA \\

\bottomrule
\end{tabular}
    \caption{Predictive performance at the level of images (bags) in the PANDA dataset. The results are the mean and standard deviation over five independent runs. The algorithm PS-MIL was run only once because of its high computational training cost.}
    \label{tab:panda_bag}
\end{table}

\section{Conclusions, limitations and future work}\label{sec:conclusions}

In this work we have introduced VGPMIL-PR-I, a novel MIL methodology that incorporates instance label correlation through a coupling term inspired by the Ising model. 
VGPMIL-PR-I is a generalization of another probabilistic MIL method, whose formulation is theoretically recovered when the influence of the Ising term converges to zero.
In the experimental section, we have shown that VGPMIL-PR-I outperforms other related state-of-the-art probabilistic MIL approaches in two real-world problems of prostate cancer detection, effectively reducing false positive bag predictions and providing instance-level predictions. We have also provided different visualizations to better understand the behavior of the proposed model, specially the influence of the new coupling term.

As discussed along the paper, our model presents several limitations which we summarize next.
Firstly, we needed to introduce a diagonal approximation to compute the expectation of the truncated mulivariate Gaussian in VGPMIL-PR-I, recall Section \ref{sec:inference}. This is probably reflected in the empirical performance, as the improvement when moving from VGPMIL to VGPMIL-PR is generally larger than when moving from VGPMIL-PR to VGPMIL-PR-I. 
Secondly, we have observed that the behaviour of VGPMIL-PR-I depends on the value of $\lambda$, which regulates the strength of the coupling term. 
Although we have discussed the role of $\lambda$ and tested different values, it remains a hyperparameter that has to be found empirically using the validation set.
We believe that its value (or even distribution over it) could be estimated from the data by introducing $\lambda$ in the probabilistic modeling. Even more, $\lambda$ could be estimated per image, since the level of correlation could be image-dependent.
Thirdly, we have observed that the image-level performance of VGPMIL-PR-I is not generally better than that of attention-based methods. This is probably due to the different nature of the models. Indeed, the explicit modeling of instance label in GP-based models, which allows them to provide instance-level predictions, may come at the cost of less accurate bag-level predictions.

In addition to the aforementioned ideas, this work opens other future research lines. First, seeing the performance boost obtained in GP-based methods through the novel coupling term, and taking into account the good results of attention-based methods in bag-level prediction, it is very interesting to explore the modeling of instance label correlations in the context of attention-based methods. 
Second, notice that we are using mean-field variational inference to estimate the model parameters in VGPMIL-PR-I. A promising alternative is to estimate them by directly optimizing the evidence lower bound (ELBO). 
Finally, although we have focused on modeling correlation between neighboring patches in histopathogical images, we expect that the ideas behind our proposal can boost further research in MIL, by exploiting the particular structure of the data used in different applications.

\section*{Acknowledgements}
This work has received funding from the European Union’s Horizon 2020 research and innovation programme under the Marie Skłodowska Curie grant agreement No 860627 (CLARIFY Project), from the Spanish  Ministry  of  Science  and Innovation under project PID2019-105142RB-C22, and by the University of Granada and FEDER/Junta de Andalucía under project B-TIC-324-UGR20 (Proyectos de I+D+i en el marco del Programa Operativo FEDER Andalucía).
PMA has been supported by the Margarita Salas postdoctoral fellowship (Spanish Ministry of Universities with Next Generation EU funds). 

\bibliography{biblio}

\end{document}